# Modeling and Simulation of Robotic Finger Powered by Nylon Artificial Muscles- Equations with Simulink model

**Lokesh Saharan, Lianjun Wu and Yonas Tadesse,**
Humanoid, Biorobotics and Smart Systems (HBS) Laboratory,
Department of Mechanical Engineering
The University of Texas at Dallas
LS: lokeshbsaharan@gmail.com, LW: lxw132630@utdallas.edu, YT: yonas.tadesse@utdallas.edu

**Abstract**

*This paper shows a detailed modeling of three-link robotic finger that is actuated by nylon artificial muscles and a simulink model that can be used for numerical study of a robotic finger. The robotic hand prototype was recently demonstrated in recent publication Wu, L., Jung de Andrade, M., Saharan, L.,Rome, R., Baughman, R., and Tadesse, Y., 2017, "Compact and Low-cost Humanoid Hand Powered by Nylon Artificial Muscles," Bioinspiration & Biomimetics, 12 (2)[1]. The robotic hand is a 3D printed, lightweight and compact hand actuated by silver-coated nylon muscles, often called Twisted and coiled Polymer (TCP) muscles. TCP muscles are thermal actuators that contract when they are heated and they are getting attention for application in robotics. The purpose of this paper is to demonstrate the modeling equations that were derived based on Euler –Lagrangian approach that is suitable for implementation in simulink model.*

### The Following is the modeling equations for 3-link robotic finger

The schematic diagram of the biomimetic finger is shown in Fig. 1 consisting of three phalanges, which correspond to the proximal, middle and distal phalanges. We have used the Euler-Lagrangian approach for the dynamic modeling of the index finger to determine the velocity Jacobians. The offset 'e' for the tendon is assumed to be a constant value of 4.5 mm. The modeling approach follows the one described by Spong et al. [2].

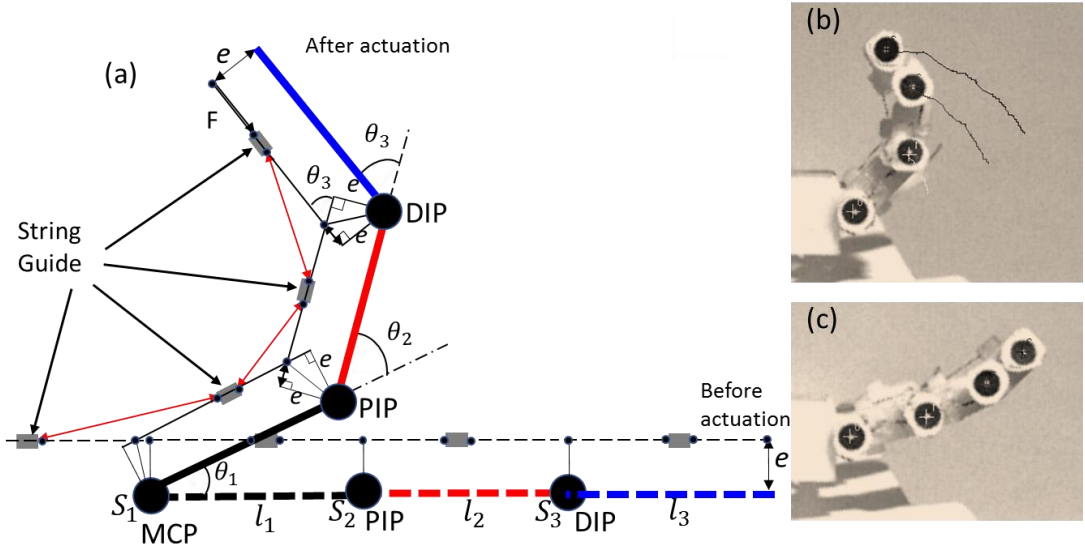

**Figure 1: (a) Free body diagram of the finger, (b) the prototype hand after actuation and (c) prototype hand before actuation.**

The following assumptions are made for the modeling of the finger dynamics: (1) the friction between the links is negligible, (2) tendon movement is smooth and experiences no jerk while passing through guides (i.e. tension is uniform throughout the string tendon), (3) all the springs have the same properties, and (4) the input force and temperature profiles of the actuator (TCP muscle) are known prior from experimental results. The general equation for the Euler-Lagrangian modeling is as follows:

$$\frac{d}{dt}\left(\frac{\partial K}{\partial \dot{q}_i}\right) - \left(\frac{\partial K}{\partial q_i}\right) + \frac{\partial P}{\partial q_i} = \tau_i \qquad (1)$$

Where: K, P and $\tau_i$ are kinetic energy, potential energy and torque of the $i^{th}$ joint respectively.

After a series of mathematical operations and transformations (Eq.(1)) can be written as:

$$\sum_j d_{kj}\ddot{q}_j + \sum_{i,j}\left\{\frac{\partial d_{kj}}{\partial q_i} - \frac{1}{2}\frac{\partial d_{ij}}{\partial q_k}\right\}\dot{q}_i\dot{q}_j + \frac{\partial P}{\partial q_k} = \tau_k \qquad (2)$$

This equation is expanded for three links and yields the following equation set:

$$\tau_1 = d_{11}\ddot{q}_1 + d_{12}\ddot{q}_2 + d_{13}\ddot{q}_3 + C_{111}\dot{q}_1^2 + C_{221}\dot{q}_2^2 + C_{331}\dot{q}_3^2 + \qquad (3a)$$

$$(C_{211} + C_{121})\dot{q}_1\dot{q}_2 + (C_{311} + C_{131})\dot{q}_3\dot{q}_1 + (C_{321} + C_{231})\dot{q}_2\dot{q}_3 + \emptyset_1 + \tau_{d1}$$

$$\tau_2 = d_{21}\ddot{q}_1 + d_{22}\ddot{q}_2 + d_{23}\ddot{q}_3 + C_{112}\dot{q}_1^2 + C_{222}\dot{q}_2^2 + C_{332}\dot{q}_3^2 + \qquad (3b)$$

$$(C_{212} + C_{122})\dot{q}_1\dot{q}_2 + (C_{312} + C_{132})\dot{q}_3\dot{q}_1 + (C_{322} + C_{232})\dot{q}_2\dot{q}_3 + \emptyset_2 + \tau_{d2}$$

$$\tau_3 = d_{31}\ddot{q}_1 + d_{32}\ddot{q}_2 + d_{33}\ddot{q}_3 + C_{113}\dot{q}_1^2 + C_{223}\dot{q}_2^2 + C_{333}\dot{q}_3^2 + \qquad (3c)$$

$$(C_{213} + C_{123})\dot{q}_1\dot{q}_2 + (C_{313} + C_{133})\dot{q}_3\dot{q}_1 + (C_{323} + C_{233})\dot{q}_2\dot{q}_3 + \emptyset_3 + \tau_{d3}$$

Where: $\tau_k$ = Torques experience by the link k, $d_{kj}$ = inertia matrix,
$\ddot{q}$ = Angular acceleration, $\dot{q}$ = Angular velocity, q = Angular Displacement,
C = Coriolis Component of acceleration,     P = Potential Energy including gravity

The impoartant equation for modeling of the finger is Eq.(3), the derivation of the Eq. (3) and Simulink® model are shown in section A1 and A2. We have ignored the terms in the model which do not have significant effects (very small in order, ~$10^{-10}$) such as Coriolis terms and all the terms of inertia matrix except the diagonal terms. But in the Eq. (3), we have added one extra damping term i.e. $\tau_{di} = c_d\,\dot{q}_i$ (i= 1,2,3) which is proportional to the angular velocity of the link $\dot{q}_i$. A similar modification of dynamic equation was presented by Lewis *et al.* [3]. Damping exists in human fingers as well as bio-mimetic robotic finger joints, which helps control (reduces oscillations).

Assuming the torque generated by the TCP actuator is distributed at each joint ($\tau_1$ = MCP joint, $\tau_2$ = PIP joint and $\tau_3$ = DIP joint) with certain factors, the torques can be written as in Eq. (4):

$$\tau_3 = \gamma\tau; \quad \tau_2 = \beta\tau; \quad \tau_1 = \alpha\tau; \quad \tau = Fe \tag{4}$$

Where $F$ is the force generated by the TCP actuator and $e$ is the offset distance of the tendon. α, β, and γ are the fractions of torque on each joint. These parameters can be determined based on the design and configuration of the finger. Zollo *et al.* [4] and Carrozza [5] have used similar assumptions to determine the applied torque at each joint. Also, the friction in the force transmission is almost negligible for such system as described in one of our previous study, which is found to be around 0.14 N when compared to the 3N produced by TCP muscle.

**A1. Euler Lagrangian Model formulation for three links**

The general equation for the Euler-Lagrangian modeling is as follows:

$$\frac{d}{dt}\left(\frac{\partial K}{\partial \dot{q}_i}\right) - \left(\frac{\partial K}{\partial q_i}\right) + \frac{\partial P}{\partial q_i} = \tau_i \tag{A1}$$

Where: K, P and $\tau_i$ are kinetic energy, potential energy and torque respectively.

The kinetic energy ($K$) has two components, the linear velocity and rotational velocity components in terms of the joint variables $q = [q_1\ q_2\ q_3]^T = [\theta_1\ \theta_2\ \theta_3]^T$ and their derivatives. The velocity terms in terms of Jacobian matrix can be described by:

$$v_i = J_{vi}(q)\dot{q}, \quad \omega_i = J_{\omega i}(q)\dot{q} \tag{A2}$$

The velocity Jacobians provided below is derived from the geometry of the links as shown in Fig. 1. The $J_{vi}$ are the velocity Jacobians ($i = 1,2,3$), which are extensions of two links velocity Jacobians described in [2, 6]. The $J_{vi}(q)$ is the Jacobian that correlates the velocity of the center of mass to the joint angular positions $\dot{q}$ and $J_{\omega i}$ is the angular velocity

Jacobian of link $i$ relative to the inertial frame of reference. The velocity Jacobians are consistent with Jacobians developed by Goutam and Aw [7].

$$J_{v1} = \begin{bmatrix} -l_{c1} \sin q_1 & 0 & 0 \\ l_{c1} \cos q_1 & 0 & 0 \\ 0 & 0 & 0 \end{bmatrix} \tag{A3a}$$

$$J_{v2} = \begin{bmatrix} -l_1 \sin q_1 - l_{c2} \sin(q_1 + q_2) & -l_{c2} \sin(q_1 + q_2) & 0 \\ l_1 \cos q_1 + l_{c2} \cos(q_1 + q_2) & l_{c2} \cos(q_1 + q_2) & 0 \\ 0 & 0 & 0 \end{bmatrix} \tag{A3b}$$

$$J_{v3} = \begin{bmatrix} w_{11} & w_{12} & w_{13} \\ w_{21} & w_{22} & w_{23} \\ w_{31} & w_{32} & w_{33} \end{bmatrix} \tag{A3c}$$

Where:

$w_{11} = -l_1 \sin q_1 - l_2 \sin(q_1 + q_2) - l_{c3} \sin(q_1 + q_2 + q_3)$

$w_{12} = -l_2 \sin(q_1 + q_2) - l_{c3} \sin(q_1 + q_2 + q_3))$

$w_{13} = -l_{c3} \sin(q_1 + q_2 + q_3))$

$w_{21} = l_1 \cos q_1 + l_2 \cos(q_1 + q_2) + l_{c3} \cos(q_1 + q_2 + q_3)$

$w_{22} = l_2 \cos(q_1 + q_2) + l_{c3} \cos(q_1 + q_2 + q_3)$

$w_{23} = l_{c3} \cos(q_1 + q_2 + q_3)$

$l_{ci}$ are the center of mass of each link obtained from SolidWorks.

*All other elements of $J_{v3}$ are 0.*

After analyzing the rotational motion of the links and due to the fact that all the joints are revolute joints, the rotational Jacobian $J_{\omega i}$ are given by:

$$J_{\omega i} = \frac{1}{2}\left\{I_1 \begin{bmatrix} 1 & 0 & 0 \\ 0 & 0 & 0 \\ 0 & 0 & 0 \end{bmatrix} + I_2 \begin{bmatrix} 1 & 1 & 0 \\ 1 & 1 & 0 \\ 0 & 0 & 0 \end{bmatrix} + I_3 \begin{bmatrix} 1 & 1 & 1 \\ 1 & 1 & 1 \\ 1 & 1 & 1 \end{bmatrix}\right\} \tag{A4}$$

Therefore, the kinetic energy of the linkage system is given by:

$$K = \frac{1}{2} m_i v_i^T v_i + \frac{1}{2} \omega_i^T I \omega_i \tag{A5}$$

Where $m_i$ is the mass of each link and $I$ is the moment of inertia about the centroid of each link. Substituting Equation (A3) and Equation (A4) into Equation (A2) and using the generalized definition of kinetic energy (Equation A5) yields Equation (A6). A rotation matrix $R_i(q)$ that correlates each link is used to transform the inertias to the inertial frame of reference (The same equation as Spong *et al.* [2]).

$$K = \frac{1}{2}\dot{q}^T \sum_{i=1}^{n}[m_i J_{vi}(q)^T J_{vi}(q) + J_{\omega i}(q)^T R_i(q) I_i R_i(q)^T J_{\omega i}(q)]\dot{q} \tag{A6}$$

The total kinetic energy can be written in short form as:

$$K = \frac{1}{2}\dot{q}^T D(q)\dot{q} \tag{A7}$$

Here, $D$ is inertia matrix of the links (robotic fingers).

$$D = \sum_{i=1}^{n}[m_i J_{vi}(q)^T J_{vi}(q) + J_{\omega i}(q)^T R_i(q) I_i R_i(q)^T J_{\omega i}(q)] \tag{A8}$$

The potential energy (assuming gravity is downward in Fig. 1), $P_g$ due to gravity for the robotic finger is as follows:

$$P_g = P_1 + P_2 + P_3 \tag{A9}$$

Where:

$$P_1 = m_1 g l_{c1} \sin q_1$$

$$P_2 = m_2 g(l_1 \sin q_1 + l_{c2} \sin(q_1 + q_2))$$

$$P_3 = m_3 g(l_1 \sin q_1 + l_2 \sin(q_1 + q_2) + l_{c3} \sin(q_1 + q_2 + q_3))$$

In our robotic hands, torsional springs are used for the return motion. The elastic potential energy $P_e$ due to the springs $kt_i$ ($i = 1,2,3$) can be written as:

$$P_e = \frac{1}{2}(kt_1 q_1^2 + kt_2 q_2^2 + kt_3 q_3^2) \tag{A10}$$

Hence the total potential energy $P$ becomes:

$$P = (m_1 l_{c1} + m_2 l_1 + m_3 l_1) g \sin q_1 + (m_2 l_{c2} + m_3 l_2) g \sin(q_1 + q_2) +$$

$$m_3 g l_{c3} \sin(q_1 + q_2 + q_3) + \frac{1}{2}(kt_1 q_1^2 + kt_2 q_2^2 + kt_3 q_3^2) \tag{A11}$$

Therefore, the potential energy derivatives are:

$$\emptyset_1 = \frac{\partial P}{\partial q_1} \qquad \emptyset_2 = \frac{\partial P}{\partial q_2} \qquad \emptyset_3 = \frac{\partial P}{\partial q_3} \tag{A12}$$

Taking the respective partial derivatives (Equation (A6)) with respect to 'q' and time 't', also using potential energy derivatives into the Euler-Lagrange Equation (Equation (A1)) and rearranging yields:

$$\sum_j d_{kj} \ddot{q}_j + \sum_{i,j} \left\{ \frac{\partial d_{kj}}{\partial q_i} - \frac{1}{2} \frac{\partial d_{ij}}{\partial q_k} \right\} \dot{q}_i \dot{q}_j + \frac{\partial P}{\partial q_k} = \tau_k \tag{A13}$$

The coefficients of the centrifugal and Coriolis components ($\dot{q}_i \dot{q}_j$) are also known as Christoffel coefficients [2] given as:

$$C_{ijk} = \left\{ \frac{\partial d_{kj}}{\partial q_i} - \frac{1}{2} \frac{\partial d_{ij}}{\partial q_k} \right\} \tag{A14}$$

where $i, j$ and $k \in (1, 2, 3)$.

Equation (A13) is the same as Spong *et al*. [2], Equation 6.55. This equation is expanded for three links and yields the following equation set:

$$\tau_1 = d_{11} \ddot{q}_1 + d_{12} \ddot{q}_2 + d_{13} \ddot{q}_3 + C_{111} \dot{q}_1^2 + C_{221} \dot{q}_2^2 + C_{331} \dot{q}_3^2 + (C_{211} + C_{121}) \dot{q}_1 \dot{q}_2 + \tag{A15a}$$

$$(C_{311} + C_{131}) \dot{q}_3 \dot{q}_1 + (C_{321} + C_{231}) \dot{q}_2 \dot{q}_3 + \emptyset_1 + \tau_{d1}$$

$$\tau_2 = d_{21} \ddot{q}_1 + d_{22} \ddot{q}_2 + d_{23} \ddot{q}_3 + C_{112} \dot{q}_1^2 + C_{222} \dot{q}_2^2 + C_{332} \dot{q}_3^2 + (C_{212} + C_{122}) \dot{q}_1 \dot{q}_2 + \tag{A15b}$$

$$(C_{312} + C_{132}) \dot{q}_3 \dot{q}_1 + (C_{322} + C_{232}) \dot{q}_2 \dot{q}_3 + \emptyset_2 + \tau_{d2}$$

$$\tau_3 = d_{31} \ddot{q}_1 + d_{32} \ddot{q}_2 + d_{33} \ddot{q}_3 + C_{113} \dot{q}_1^2 + C_{223} \dot{q}_2^2 + C_{333} \dot{q}_3^2 + (C_{213} + C_{123}) \dot{q}_1 \dot{q}_2 + \tag{A15c}$$

$$(C_{313} + C_{133}) \dot{q}_3 \dot{q}_1 + (C_{323} + C_{233}) \dot{q}_2 \dot{q}_3 + \emptyset_3 + \tau_{d3}$$

But in the Equation (A15), we have added one extra damping term i.e. $\tau_{di} = c_d \dot{q}_i$ (i= 1,2,3) which is proportional to the angular velocity of the link $\dot{q}_i$. A similar modification of dynamic equation was presented by Lewis *et al.* [3].

The inertia matrix D was generated by the application of series of mathematical operations using Equation (A8). We used MATLAB symbolic tool to solve the coefficients of the inertia matrix and checked manually as well.

$$D = \begin{bmatrix} d_{11} & d_{12} & d_{13} \\ d_{21} & d_{22} & d_{23} \\ d_{31} & d_{32} & d_{33} \end{bmatrix} \tag{A16}$$

Where:

$d_{11} = m_1 l_{c1}^2 + m_2\{l_1^2 + l_{c2}^2 + 2l_1 l_{c2} \cos q_2\} + m_3\{l_1^2 + l_2^2 + l_{c3}^2 + 2l_1 l_2 \cos q_2 +$

$2l_2 l_{c3} \cos q_3 + l_{c3} \cos(q_2 + q_3)\} + I_1 + I_2 + I_3$

$d_{12} = d_{21} = m_2(l_{c2}^2 + l_1 l_{c2} \cos q_2) + m_3\{l_2^2 + l_{c3}^2 + 2l_2 l_{c3} \cos q_3 +$

$l_1 l_2 \cos q_2 + l_1 l_{c3} \cos(q_2 + q_3)\} + I_2 + I_3$

$d_{13} = d_{31} = m_3\{l_{c3}^2 + l_1 l_{c3} \cos(q_2 + q_3) + l_2 l_{c3} \cos q_3\} + I_3$  (A17)

$d_{23} = d_{32} = m_3(l_{c3}^2 + l_2 l_{c3} \cos q_3) + I_3$

$d_{22} = m_3(l_2^2 + l_{c3}^2 + 2l_2 l_{c3} \cos q_3) + m_2 l_{c2}^2 + I_2 + I_3$

$d_{33} = m_3 l_{c3}^2 + I_3$

Once the inertial matrix is found, the Christoffel coefficients ($C_{111}, C_{222}, \ldots C_{333}$), are obtained using Equation (A14) and the results are summarized as :

$C_{112} = m_2 h_1 + m_3(h_2 + h_4)$

$C_{113} = m_3(h_2 + h_3)$

$C_{123} = C_{213} = C_{223} = m_3 h_3$

$C_{121} = C_{211} = C_{221} = -m_2 h_1 - m_3(h_2 + h_4)$

$$C_{232} = C_{322} = C_{332} = C_{132} = C_{312} = -m_3 h_3$$

$$C_{131} = C_{311} = C_{231} = C_{321} = C_{331} = -m_3(h_2 + h_3) \tag{A18}$$

$$C_{111} = C_{222} = C_{122} = C_{212} = C_{133} = C_{313} = C_{233} = C_{323} = C_{333} = 0$$

Where:

$$h_1 = l_1 l_{c2} \sin q_2; \quad h_3 = l_2 l_{c3} \sin q_3; \quad h_2 = l_1 l_{c3} \sin(q_2 + q_3); \quad h_4 = l_1 l_2 \sin q_2$$

Where the $d_{ij}$ are components of the inertia matrix (matrix $D$) and $C_{ijk}$ are the coefficients of the centrifugal and Coriolis components ($\dot{q}_i \dot{q}_j$). $q_i$ is the angular displacement (the same $\theta_i$ as defined in Fig. 1), $\dot{q}_i$ is the angular velocity, $\ddot{q}_i$ is the angular acceleration, $\emptyset_i$ is the partial derivative of potential energy with respect to joint $q_i$, $\tau_{di} = c_d \dot{q}_i$ (i = 1, 2, 3) is a damping torque and $\tau_i$ is the torque at a joint. Similar equations were derived by Li *et al.* [8] for calculating the moments of the joints of three fingers in order to determine the effect of the extrinsic and intrinsic muscles on the movement of the finger. Joint friction and structural damping can be considered in the dynamic equation, as shown in ref. [9]. Damping is an integral part of the human [10, 11] as well as bio-mimetic robotic finger joints [12, 13], which helps control and complete dexterous movements. Therefore, as stated earlier, we have considered the effect of damping factor in Equation A15 by adding terms $\tau_{di}$.

**A2. Simulink Model for the Finger Joint**

Based on the equations shown in the previous section, a simulink model was created (Fig.2). It consists of three main blocks corresponding to the three links, Link 1, Link 2 and Link 3. The input for each block is the torque shown in the left side of the figure 2. Outputs from the simulations are shown in red color. Since the equations are coupled, signal routings were used to simplify the diagram. There are multiple sub-blocks that are included in the diagram, and all the details can be seen in Fig.2. Since three link robotic fingers are common, such detailed simulation diagram/equations and tools will be useful to study dynamic motion of fingers that are actuated by different means.

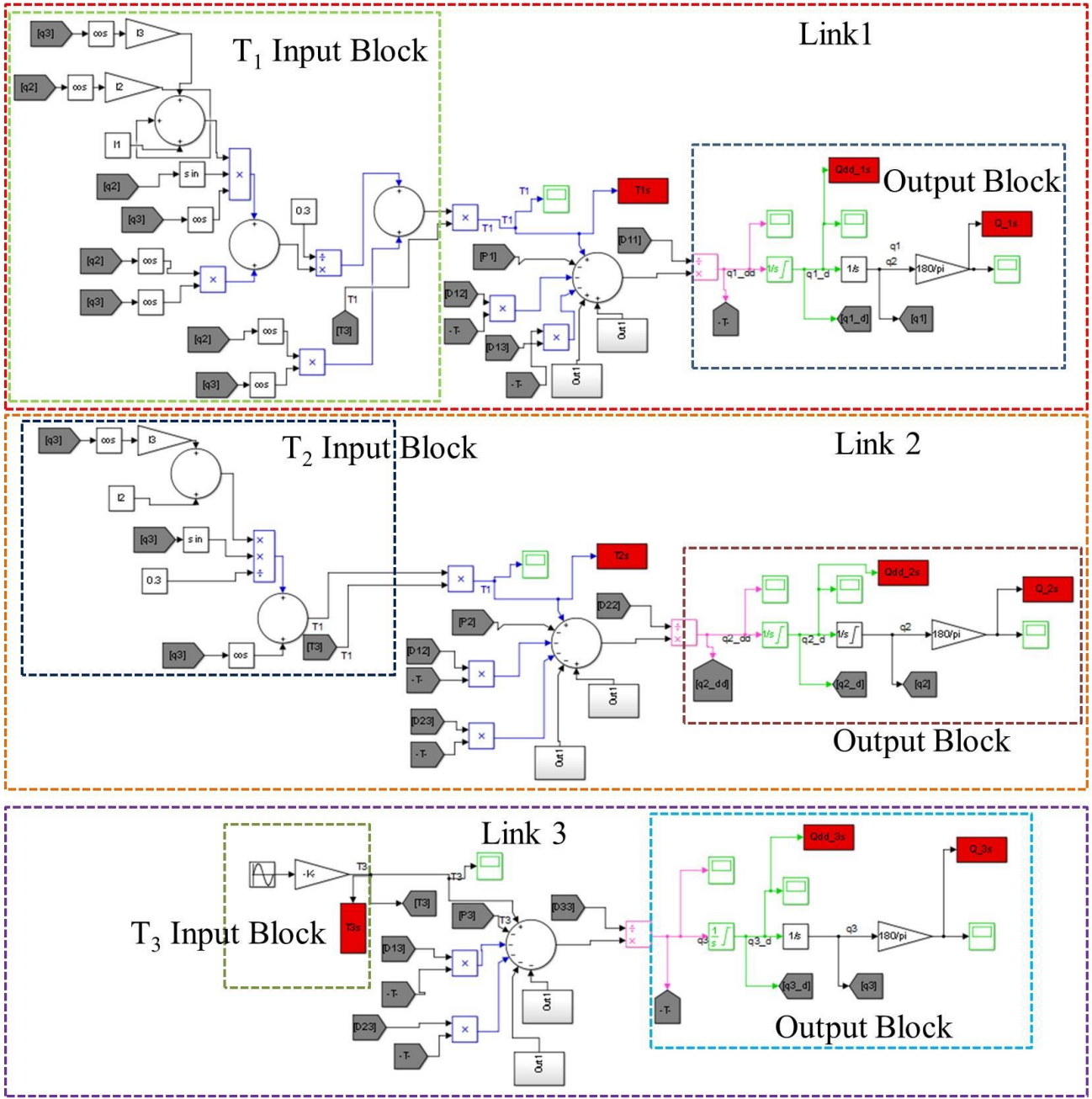

**Figure 2: Block diagram of the Simulink model used to model the nylon muscle.**